\documentclass[11pt,a4paper]{article}
\usepackage{times,latexsym}
\usepackage{url}
\usepackage[T1]{fontenc}

\usepackage[acceptedWithA]{tacl2021v1}

\usepackage{xspace,mfirstuc,tabulary}

\newif\iftaclinstructions
\taclinstructionsfalse 
\iftaclinstructions
\renewcommand{\confidential}{}
\renewcommand{\anonsubtext}{(No author info supplied here, for consistency with
TACL-submission anonymization requirements)}
\newcommand{\instr}
\fi

\iftaclpubformat 

\else

\fi

\usepackage{amsmath}
\usepackage{graphicx}
\usepackage{caption}
\usepackage{subcaption}
\usepackage{booktabs}
\usepackage{soul}

\setstcolor{red}

\title{Temporal Effects on Pre-trained Models for Language Processing Tasks}

\author{Oshin Agarwal \\
        University of Pennsylvania \\
        {\tt oagarwal@seas.upenn.edu} \\\And
        Ani Nenkova \\
        Adobe Research \\
        \texttt{nenkova@adobe.com}
}

\date{}

\begin{document}
\maketitle
\begin{abstract}

Keeping the performance of language technologies optimal as time passes is of great practical interest.
We study temporal effects on model performance on downstream language tasks, establishing a nuanced terminology for such discussion and identifying factors essential to conduct a robust study.
We present experiments for several tasks in English where the label correctness is not dependent on time and demonstrate the importance of distinguishing between temporal model deterioration and temporal domain adaptation for systems using pre-trained representations. We find that depending on the task, temporal model deterioration is not necessarily a concern. Temporal domain adaptation however is beneficial in all cases, with better performance for a given time period possible when the system is trained on temporally more recent data. Therefore, we also examine the efficacy of two approaches for temporal domain adaptation without human annotations on new data. Self-labeling shows consistent improvement and notably, for named entity recognition, leads to better temporal adaptation than even human annotations.
\end{abstract}

\section{Introduction}

Language models capture properties of language, such as semantics of words and phrases and their typical usage, as well as facts about the world expressed in the language sample on which they were trained. Effective solutions for many language tasks depend, to a varying degree, on the background knowledge encoded in language models. Performance may degrade as language and world-related facts change. In some scenarios, language will change as a result of deploying a system that uses the language to make a prediction, as in spam detection \cite{fawcett2003vivo}. But most change is not driven by such adversarial adaptations: the language expressing sentiment in product reviews \cite{lukes-sogaard-2018-sentiment}, the named entities and the contexts in which they are discussed on social media \cite{fromreide-etal-2014-crowdsourcing,rijhwani-preotiuc-pietro-2020-temporally} and language markers of political ideology \cite{huang-paul-2018-examining} all change over time.

Whether and how this change impacts the performance of different language technologies is a question of great practical interest.
Yet research on quantifying how model performance changes with time has been sporadic. Moreover, approaches to solving language tasks have evolved rapidly, from bag of words models which rely on a small number of fixed words represented as strings, without underlying meaning, to fixed dense word representations such as word2vec and GloVe \cite{Mikolov2013DistributedRO,pennington-etal-2014-glove} and large contextualized representations of language \cite{peters-etal-2018-deep,devlin-etal-2019-bert} that are trained on task-independent text to provide a backbone representation for word meaning.  The swift change in approaches has made it hard to understand how representations and the data used to train them modulate the changes in system performance over time.

We present experiments (\S \ref{sec:setup} \& \S \ref{sec:metrics}) designed to study temporal effects on downstream language tasks, disentangling worsening model performance due to temporal changes ({\em temporal model deterioration}) and the benefit from retraining systems on temporally more recent data in order to obtain optimal performance ({\em temporal domain adaptation}). We present experiments on four tasks for English---named entity recognition, truecasing, sentiment and domain classification. We work only with tasks where the correctness of the label is not influenced by time, unlike other tasks such as open domain question answering where the answer may depend on the time when the question are posed (e.g.\ who is the CEO of X?). For each task, we analyze how the performance of approaches built on pre-trained representations changes over time and how retraining on more recent data influences it (\S \ref{sec:results}). We find that models built on pre-trained representations do not experience temporal deterioration on all tasks. However, temporal domain adaptation is still possible, i.e.\ performance can be further improved by retraining on more recent human labeled data.

We further find that neural models fine-tuned on the same data but initialized with random vectors for word representation exhibit dramatic temporal deterioration on the same datasets (\S \ref{sec:random}). Models powered by pre-trained language models however are not impacted in the same way. Unlike in any prior work, we study several representations (\S \ref{sec:repr}) including those built using the same architecture and data but different model sizes (\S \ref{sec:size}).

Even though the pre-training data of several representations overlaps in time with task-specific data and some confounding is possible, two sets of experiments show that it is unlikely (\S \ref{sec:time}).
These results  provide strong evidence for model deterioration without pre-training; it also raises questions for future work, on how the (mis)match between task data and pre-training data influences performance, with greater mismatch likely to be more similar to random initialization, resulting in a system more vulnerable to temporal deterioration.

The central insight from our work is that performance of pre-trained models on downstream tasks where answer correctness is time-independent, does not necessarily deteriorate over time but that the best performance at a given time can be obtained by retraining the system on more recent data. Furthermore, based on the experiments to assess the impact of different components of a model, we provide recommendations for the design of future studies on temporal effects (\S \ref{sec:recommend}). This will make it both easier to conduct future studies and have more robust findings by controlling confounding factors and ignoring others.

Finally, we present two methods for temporal adaptation that do not require manual labeling over time (\S \ref{sec:temporal_adapt}).
One of the approaches is based on continual pre-training where we modify the typical domain adaptative pre-training with an additional step. The second method relies on self-labeling and is highly effective with consistent improvement across all settings. On one of the datasets, self-labeling is even superior to fine-tuning on new labeled human annotated data.

\section{Background and Related Work}

Language changes over time \cite{Weinreich1968EmpiricalFF,eisenstein-2019-measuring,mcculloch2020because}. For longer time periods, a robust body of computational work has proposed methods for modeling the changes in active vocabulary \cite{Dury2011WhenTD,DBLP:conf/www/Danescu-Niculescu-MizilWJLP13} and in meaning of words \cite{10.1145/2064448.2064475,hamilton-etal-2016-diachronic,rosenfeld-erk-2018-deep,brandl-lassner-2019-times}. Changes in vocabulary and syntax, approximated by bi-grams in \citet{eisenstein-2013-bad}, also occur on smaller time scales, such as days and weeks, and occur more in certain domains, e.g.\ change is faster in social media than in printed news. Such language changes over time can also be approximated by the change in language model perplexity. \citet{Lazaridou2021PitfallsOS} find that language model perplexity changes faster for politics and sports than for other domains, suggesting that these domains evolve faster than others. They also demonstrate that language models do not represent well language drawn from sources published after it was trained: perplexity for text samples drawn from increasingly temporally distant sources increases steadily.  Their qualitative analysis shows that the changes are not only a matter of new vocabulary: even the context in which words are used changes.

The global changes captured with language model perplexity and analysis of individual words cannot indicate how these changes impact the performance of a model for a given task. \citet{Rttger2021TemporalAO} present a meticulously executed study of how domain change (topic of discussion) influences both language models and a downstream classification task. They show that even big changes in language model perplexity may lead to small changes in downstream task performance. They also show that domain adaptation and temporal adaptation are both helpful for the downstream classification task they study, with domain adaptation providing the larger benefit.

Here, we also focus on the question of how time impacts downstream tasks. Studying temporal change in model performance requires extra care in experimental design to tease apart the temporal aspect from all other changes that may occur between two samples of testing data. Teasing apart temporal change from domain change is hardly possible. Even data drawn from the same source may include different domains over time\footnote{\citet{huang-paul-2018-examining} find the topics in their data and observe that the top 20 topics change over time.}.
Despite these difficulties, there are two clear and independent questions that we pursue, related to system performance over time.

\subsection{Does performance deteriorate over time?}

To study this question of {\em temporal model deterioration}, we need to measure performance over several time periods.
Let $d_s$, $d_t$ and $d_n$ denote respectively the first, $t^{th}$ ($s \le t \le n$) and last temporal split in a dataset.
To guard against spurious conclusions that reflect specifics of data collected in a time period, the starting point for the analysis should also vary. \citet{huang-paul-2018-examining} use such a set up, performing an extensive evaluation by training $n$ models on $d_s$ to $d_n$ and then evaluating them on all remaining $n-1$ splits, both on past and future time periods. The resulting tables are cumbersome to analyze but give a realistic impression of the trends. We adopt a similar setup for our work, reporting results for a number of tasks with models trained on data from different time periods and tested on data from all subsequent time periods available. In addition, we introduce summary statistics that capture changes across all studied time periods to compare the temporal trends easily and to compute statistical significance for the observed changes in performance (\S \ref{sec:setup} \& \ref{sec:metrics}).

Most prior work, in contrast, uses a reduced setup \cite{lukes-sogaard-2018-sentiment, rijhwani-preotiuc-pietro-2020-temporally, sogaard-etal-2021-need} with a fixed test time period and measures the performance of models trained on different time periods on this fixed test set.
Such  evaluation on one future temporal split does not measure the change in model performance over time and cannot support any conclusions about temporal deterioration.\footnote{\citet{Lazaridou2021PitfallsOS} omit such an evaluation since they measure language model perplexity which is sensitive to document length, which they found differed across months. \citet{Rttger2021TemporalAO} evaluate over multiple test sets on a classification task but also omit such an evaluation by reporting the change in the metrics of models w.r.t.\ a control model without temporal adaptation.} This setup from prior work supports conclusions only about temporal domain adaptation i.e.\ whether retraining on temporally new data helps improve performance on future data, with a single point estimate for the improvement.

\subsection{Can performance at time $t$ be improved?}
As described above, most prior work chose $d_n$, the data from latest time period as the test data, to evaluate models trained on earlier data. \citet{lukes-sogaard-2018-sentiment} train a model for sentiment analysis of product reviews in 2001-2004 and 2008-2011 and test them on reviews from 2012-2015.
\citet{rijhwani-preotiuc-pietro-2020-temporally} train models for named entity recognition on tweets from each year from 2014 to 2018 and test them on tweets from 2019.
\citet{sogaard-etal-2021-need} work with the tasks of headline generation and emoji prediction. For headline generation, they successively train models on data from 1993 to 2003 and test it on data from 2004. For emoji prediction, the training data comes from different days and the last one is used as the test set.
\citet{Lazaridou2021PitfallsOS} train a language model on various corpora with test data from 2018-2019 and train years that either overlap with the test year or precede them.

Such results allow us to draw conclusions about the potential for {\em temporal domain adaptation}, revealing that models trained on data closer to the test year perform better on that test year. The only problem is that there is a single test year chosen and any anomaly in that test year may lead to misleading results. The temporal Twitter corpus \cite{rijhwani-preotiuc-pietro-2020-temporally}, where 2019 is the dedicated test year, is an instructive case in point. Twitter increased the character limit in late 2017. As a result, tweets from 2018 are longer and contain more entities than these in prior years. The potential for temporal adaptation measured only on 2018 data contrasted with prior years may give a highly optimistic view for how much models can improve. An evaluation setup like the one in \citet{huang-paul-2018-examining} or the recent work in \citet{Rttger2021TemporalAO} is needed to draw robust conclusions. We adopt their setup with some changes. We also introduce summary statistics to easily interpret trends and a test for significance to determine if the changes in performance are compatible with random fluctuation of performance across time periods.

Another line of work on temporal effects focuses on temporal adaptation by incorporating time in the training process as opposed to retraining models on new human labeled data regularly. Several approaches have been proposed such as diachronic word embeddings, the ``frustratingly simple'' domain adaptation, aligning representations of old and new data, time-aware self-attention and continual learning as new data is available \cite{10.5555/3304889.3304971,huang-paul-2019-neural-temporality,Bjerva2020BackTT,hofmann-etal-2021-dynamic,rosin2022temporal}. An expanded evaluation of these approaches to measure deterioration and adaptation across several time periods with different representations will be useful, given our findings.

\section{Experimental Resources}

Here we describe the datasets and the different models used.

\subsection{Tasks and Datasets}
\label{sec:datasets}

We use four English datasets, two for sequence labeling and two for text classification\footnote{More dataset details and model hyperparameters can be found in the appendix and at \url{https://github.com/oagarwal/temporal-effects}}.

\paragraph{Named Entity Recognition with Temporal Twitter Corpus} TTC \cite{rijhwani-preotiuc-pietro-2020-temporally} consists of tweets annotated with {\em PER}, {\em LOC} and {\em ORG} entities. There are 2,000 tweets in each year from the period 2014--2019. TTC is the only corpus with human annotations specifically collected in order to study temporal effects on performance. Other datasets, including the three we describe next are in fact derived annotations that do not require manual annotation.

\paragraph{Truecasing with New York Times} Truecasing \cite{Gale1995DiscriminationDF, lita-etal-2003-truecasing} is the task of case restoration in text. We sample a dataset from the NYT Annotated Corpus \cite{sandhaus2008new} which has sentences that follow English orthographic conventions. We perform a constrained random sampling of 10,000 sentences per year from 1987--2004 and organize the data with three consecutive years per split. To maintain diversity of text, we select an approximately equal number of sentences from each domain (indicated by the metadata) and only two sentences per article. Sentences should have at least one capitalized word, not including the first word and should not be completely in uppercase (headlines appear in all uppercase). We model the task as sequence labeling with binary word labels of fully lowercase or not.

\paragraph{Sentiment Classification with Amazon Reviews} AR \cite{ni-etal-2019-justifying} consists of 233M product reviews rated on a scale of 1 to 5. Following prior work \cite{lukes-sogaard-2018-sentiment}, we model this task as binary classification, treating rating of greater than 3 as positive and the remaining as negative. We randomly sample 40,000 reviews per year from the period 2001--2018 and organize the data with three consecutive years per split. The first 50 words of each review are used.

\paragraph{Domain Classification with New York Times} We select the first 40,000 articles from each year in 1987--2004 from the NYT Annotated Corpus and organize the data with three consecutive years per split. The article domain is labeled using the article metadata. Certain domains are merged based on the name overlap, resulting in eight domain---Arts, Business, Editorial, Financial, Metropolitan, National, Sports and Others. The first 50 words (1-2 paragraphs) of each article are used.

\subsection{Models}
\label{sec:models}

We use two architectures (biLSTM-CRF and Transformers) and four representations (GloVe, ELMo, BERT, RoBERTa) for the experiments. Hyperparameters and other fine-tuning details are noted in the appendix.

\paragraph{GloVe+char} BiLSTM \cite{hochreiter1997long} with 840B-300d-cased GloVe \cite{pennington-etal-2014-glove} and character-based word representation \cite{ma-hovy-2016-end} as input. For sequence labeling, a CRF \cite{lafferty2001conditional} layer is added and prediction is made for each word. For text classification, the representation of the first word is used to make the prediction.

\paragraph{ELMo+GloVe+char\footnote{This combination yields better results than ELMo alone.}} Same as GloVe+char but
the Original
ELMo \cite{peters-etal-2018-deep} embeddings are concatenated to the input.

\paragraph{BERT} \cite{devlin-etal-2019-bert} We use the large model for sequence labeling and the base model for text classification, both cased. The number of training examples was larger for text classification resulting in a much faster base model with minimally lower performance than the large one.

\paragraph{RoBERTa} \cite{Liu2019RoBERTaAR} We use the large model for sequence labeling and the base model for text classification.

\section{Experimental Setup}
\label{sec:setup}

We divide each dataset into $n$ temporal splits with equal number of sentences for sequence labeling and equal number of documents for text classification to minimize any performance difference due to the size of the split. We randomly downsample to the size of the smallest temporal split whenever necessary. Let $d_s$, $d_t$ and $d_n$ denote the first, $t^{th}$ and last temporal split in the dataset respectively.

\paragraph{Train and Test Set}
We largely follow \citet{huang-paul-2018-examining}, with minor clarifications on certain aspects as well as additional constraints due to difference in dataset size across tasks, ensuring consistency in setup.
First, we vary both training and test year but limit the evaluation to future years since we want to mimic the practical setup of model deployment. We train $n-1$ models, each on a specific temporal split, starting from a model on $d_s$ to a model on $d_{n-1}$, and evaluate the model trained on $d_t$ on test sets starting from $d_{t+1}$ to $d_n$. Each temporal split has the same number of sentences/documents and training/evaluation is done only on data from a given split (not cumulative data). Increase in training data size is typically associated with increase in performance, so cumulative expansion of the training set will introduce a confound between the temporal effects and dataset size. With these results, a lower triangular matrix can be created with the training years as the columns and the test years as the rows. A sample can be seen in Table~\ref{table:drift_ttc}.

Next, we need to further divide each temporal split $d_t$ into three sub-splits for training, development and testing. We are limited by our smallest dataset on NER, which is by far the hardest to label and is the only task that requires manual data annotation. It has 2,000 sentences in each year and splitting it into three parts will not provide us with enough data to train a large neural model or reliably evaluate it. Hence, we do not evaluate on the current year but only on the future ones. When training a model on $d_t$, it is split 80-20 into a training set $train_t$ and a development set $dev_t$. Both these sets combined i.e.\ the full $d_t$ serves as the test set $test_t$ when a past model is evaluated on it.

\paragraph{Development Set}
The model checkpoint that performs best on the development set is typically chosen as the model to be tested.
Yet prior work either does not report full details of the data used for choosing hyperparameters, or uses default hyperparameters, or draws the development set from the same year as the test set year
\cite{fromreide-etal-2014-crowdsourcing, lukes-sogaard-2018-sentiment, rijhwani-preotiuc-pietro-2020-temporally, Chen2021MitigatingTA, sogaard-etal-2021-need}.
We choose development data from the time period of the training data, reserving 20\% of the data in each temporal split, since data from a future time period will not be available to use as the development set during training. Beyond concerns about setup feasibility,
through experiments not presented in detail because of space constraints,
we found that the selection of development set from the test year may affect performance trends and even lead to exaggerated improvement for temporal domain adaptation.

\section{Evaluation Metrics}
\label{sec:metrics}

\paragraph{Task Metrics}
In the full matrix described above, we report task-specific metrics, by averaging them over three runs with different random seeds. For NER, we report the span-level micro-F1 over all entity classes; for truecasing, we report F1 for the cased class. For sentiment classification, we report F1 for the negative sentiment class; for domain classification, we report the macro-F1 over all the domains. The positive sentiment and uncased word account for about 80\% of the data in their respective tasks and are largely (but not completely) unaffected over time.

\paragraph{Temporal Summary Metrics} For a compact representation, we also report summary deterioration score (DS) and summary adaptation score (AS) in addition to the full matrix with the task-specific evaluation results. Deterioration score measures the average change in the performance of a model over time. A negative score indicates that the performance has deteriorated. Similarly, the adaptation score measures the average improvement in performance by retraining on recent data, labeled or unlabeled (\S \ref{sec:temporal_adapt}). A positive score means performance improves by retraining.

For each score, we report two versions, one that measures the average change between immediately consecutive time periods and the other that measures the change with respect to an anchor (oldest) time period since retraining need not be at regular intervals. The anchor-based scores are also a more stable metric since the amount of time passed between the values being compared is longer and therefore we are more likely to observe discernible temporal effects. For measuring deterioration, the anchor is the oldest test time period for the given model i.e.\ if a model is trained on $d_t$, then the task metric on $d_{t+1}$ is the anchor score (first available row in each column of the full results matrix). For measuring adaptation, the anchor is the oldest train time period so the anchor is always $d_s$ (first column in the full results matrix). Let $M_i^j$ be the task metric measured on $d_j$ when the model is trained on $d_i$. Let N be the number of elements in the sum and $d_a$ be the anchor time period.  The summary scores are defined as follows.

\begin{gather*}
    DS_{t}^{t-1} = \frac{1}{N} \sum_{i \in train} \sum_{j \in test}  M_i^{j+1} - M_i^j
\end{gather*}

\begin{gather*}
    DS_{t}^{a} = \frac{1}{N} \sum_{i \in train} \sum_{j \in test}  M_i^{j+1} - M_i^a
\end{gather*}

\begin{gather*}
    AS_{t}^{t-1} = \frac{1}{N} \sum_{i \in train} \sum_{j \in test}  M_{i+1}^j - M_i^j
\end{gather*}

\begin{gather*}
    AS_{t}^{a} = \frac{1}{N} \sum_{i \in train} \sum_{j \in test}  M_{i+1}^{j} - M_a^j
\end{gather*}

To test if a given trend for deterioration or adaptation is statistically significant, we consider the vector of differences in each of the formulae above, and run a two-sided Wilcoxon signed-rank test to check if the median difference is significantly different from zero. For our setup there are 10 differences total, corresponding to a sample size of $N$=10. When we report deterioration and adaptation scores in tables with results, we indicate with an asterisk (*) values corresponding to a vector of differences with p-value smaller than 0.05. While this measures the fluctuations across the average task-metrics over different training and test years, it does not take into account the variations across different runs of the same model with random seeds. An ideal test would take into account both the random seeds and the different train/test years. However, this is not straightforward and we leave the design of such a test for future work. Instead, in this paper, to ensure trends are not affected by variations across seeds, we calculate three values for each of the four scores, corresponding to the three runs. For deterioration, the performance of a model trained with a specific seed is measured over time, but for adaptation, the performance change may be measured w.r.t.\ a model trained with a different seed, as will be the case in practice. We then report the minimum and maximum of this score for the significant summary metrics as measured above. If the sign of the minimum and maximum of each score is the same, the trend in the scores remains same across runs, even if the magnitude varies.

Along with the summary scores, we also report three salient values of the task metric from the full results table (Table~\ref{table:drift_ttc}) in the summary ($M_s^{s+1}$, $M_s^n$ and $M_{n-1}^n$), necessary to compare the relative performance across datasets and representations. Remember that $M_i^j$ is the evaluation metric measured on $d_j$ when the model is trained on data split $d_i$. $M_s^{s+1}$, which is the value in the first row and first column in the full results table, represents the task metric when the model is trained on the first temporal split and evaluated on the immediate next one. It serves as the base value for comparison. $M_s^n$, which is the value in the last row and first column in the full results table, shows whether the performance of the model deteriorated from $M_s^{s+1}$ over the longest time span available in the dataset, by comparing the performance of the same model on the last temporal split. Similarly, $M_{n-1}^n$, which is the value in the last row and last column of the full results table, shows if the performance can be improved by retraining from $M_s^n$ over the longest time span available in the dataset, by retraining on the latest available temporal split.

\section{Main Results}
\label{sec:results}

Results are shown in Table~\ref{table:drift_ttc} and Table~\ref{table:summary_change_short}. For NER, we show the full matrix with the task metrics but for all other tasks, we only report the summary scores. Here, we only report results with the oldest (GloVe) and latest (RoBERTa) representation used in our experiments. For other representations, we provide a detailed analysis in later sections.

\begin{table}[t]
\centering
\footnotesize
\begin{tabular}{*6c} \toprule
& \multicolumn{5}{c}{Train Year}\\ \cmidrule(lr){2-6}
Test Year & 2014 & 2015 & 2016 & 2017 & 2018\\ \midrule
\multicolumn{6}{c}{GloVe+char biLSTM-CRF}\\ \midrule
2015 & 55.18 & -     & -     & -     & -     \\
2016 & 56.22 & 57.13 & -     & -     & -     \\
2017 & 55.09 & 53.95 & 59.43 & -     & -     \\
2018 & 51.06 & 53.12 & 57.75 & 57.82 & -     \\
2019 & 54.10 & 54.56 & 59.48 & 60.41 & 62.99 \\ \midrule
\multicolumn{6}{c}{RoBERTa}\\ \midrule
2015 & 67.48 & -     & -     & -     & -     \\
2016 & 69.41 & 72.02 & -     & -     & -     \\
2017 & 68.30 & 70.53 & 70.29 & -     & -     \\
2018 & 67.82 & 68.33 & 69.29 & 68.60 & -     \\
2019 & 77.79 & 78.33 & 78.89 & 78.28 & 79.99\\ \bottomrule
\end{tabular} 
\caption{F1 for NER on TTC. Training is on gold-standard data.}
\label{table:drift_ttc}
\end{table}
\begin{table}[t]
\centering
\footnotesize
\begin{tabular}{lc@{\hspace{0.25em}}c@{\hspace{0.25em}}cc@{\hspace{0.25em}}c@{\hspace{0.25em}}c@{\hspace{0.25em}}c} \toprule
& $M_s^{s+1}$ & $M_s^n$ & $M_{n-1}^n$ & $D_{t}^{a}$ & $A_{t}^{a}$ & $D_{t}^{t-1}$ & $A_{t}^{t-1}$\\ \cmidrule(lr){2-4}\cmidrule(lr){5-8}

\multicolumn{8}{c}{NER-TTC}\\ \midrule
GloVe & 55.2 & 54.1 & 63.0 & -1.3$^{\hphantom{*}}$ & 4.1$^{*}$ & -0.1$^{\hphantom{*}}$ & 2.1$^{*}$\\
RoBERTa & 67.5 & 77.8 & 80.0 & 3.2$^{\hphantom{*}}$ & 1.4$^{*}$ & 3.5$^{\hphantom{*}}$ & 0.8$^{\hphantom{*}}$ \\ \midrule

\multicolumn{8}{c}{Truecasing-NYT}\\ \midrule
GloVe & 93.8 & 93.0 & 94.6 & -0.6$^{*}$ & 0.3$^{\hphantom{*}}$ & -0.2$^{*}$ & 0.3$^{\hphantom{*}}$\\
RoBERTa & 97.5 & 94.4 & 95.6 & -1.1$^{\hphantom{*}}$ & 0.4$^{*}$ & -0.8$^{\hphantom{*}}$ & 0.2$^{*}$ \\ \midrule

\multicolumn{8}{c}{Sentiment-AR}\\ \midrule
GloVe & 44.9 & 42.8 & 64.7 & 0.8$^{\hphantom{*}}$ & 10.3$^{*}$ & 0.4$^{\hphantom{*}}$ & 4.9$^{*}$\\
RoBERTa & 69.9 & 73.9 & 78.9 & 2.5$^{*}$ & 2.5$^{*}$ & 1.3$^{*}$ & 1.1$^{*}$\\ \midrule

\multicolumn{8}{c}{Domain-NYT}\\ \midrule
GloVe & 73.0 & 68.4 & 78.1 & -2.7$^{*}$ & 7.9$^{*}$ & -0.5$^{\hphantom{*}}$ & 3.6$^{*}$\\
RoBERTa & 84.2 & 78.2 & 86.6 & -3.7$^{*}$ & 5.8$^{*}$ & -1.1$^{*}$ & 2.9$^{*}$\\ \bottomrule

\end{tabular} 
\caption{Deterioration and Adaptation scores for models fine-tuned on gold standard data. Positive and negative scores denotes an increase and decrease in the task metric respectively. An asterisk marks statistically significant scores.}
\label{table:summary_change_short}
\end{table}

\paragraph{Temporal Model Deterioration} can be tracked over the columns in the full matrix and by comparing $M_s^{s+1}$ and $M_s^{n}$ along with the deterioration scores in the summary table. Each column in the full matrix presents the performance of a fixed model over time on future data. We do not observe temporal deterioration for all cases. For NER, we observe deterioration with GloVe but not with RoBERTa, for which performance improves over time. However, neither of the deterioration scores are statistically significant. For sentiment, there is no deterioration; in fact model performance improves over time (significant for RoBERTa). For truecasing, there is some deterioration (significant for GloVe). For domain classification, there is considerable deterioration (significant for both representations). The difference between the two versions of the deterioration scores is as expected, smaller for consecutive periods and larger when computed with respect to the anchor.

Model deterioration appears to be both task and representation dependent. This result offers a contrast to the findings in \citet{Lazaridou2021PitfallsOS} that language models get increasingly worse at predicting future utterances. We find that not all tasks suffer from model deterioration. The temporal change in vocabulary and facts does not affect all tasks as these changes and information might not be necessary to solve all tasks. These results do not depend on whether pre-training data and task data overlap temporally (\S \ref{sec:time}).

\paragraph{Temporal Domain Adaptation} can be tracked over the rows in the full matrix and by comparing $M_s^{n}$ and $M_{n-1}^{n}$ along with the adaptation scores in the summary table. Each row in the full matrix represents performance on a fixed test set starting with models trained on data farthest away to the temporally nearest data. Performance improves with statistical significance as the models are retrained on data that is temporally closer to the test year. The results are consistent with prior work that uses non-neural models \cite{fromreide-etal-2014-crowdsourcing,lukes-sogaard-2018-sentiment,huang-paul-2018-examining} or evaluates on a single test set \cite{lukes-sogaard-2018-sentiment, rijhwani-preotiuc-pietro-2020-temporally, sogaard-etal-2021-need}. However, the extent of improvement varies considerably by test year, task and representation. The largest improvement is for the domain classification followed by the sentiment classification. It is worth noting that both of these datasets span 18 years whereas the NER dataset spans 6 years and more improvement may be observed for NER for a similar larger time gap. The change in performance on truecasing is almost non-existent. The difference between the two versions of the adaptation scores is as expected given the longer gap between retraining.

For all four summary scores, we also report the minimum and maximum by calculating three values of each score corresponding to three different runs (\S \ref{sec:metrics}). The results are shown in the appendix. While the extent of deterioration and adaptation varies across runs, the sign of the scores is the same for the maximum and minimum, i.e.\ the trends are consistent across runs.

\section{No Pre-training}
\label{sec:random}

Above we found that models powered by pre-trained representations do not necessarily manifest temporal deterioration. At first glance, our findings may appear to contradict findings from prior work. They appear more compatible though when we note that most of the early work discussing temporal effects on model performance studied bag of words models \cite{fromreide-etal-2014-crowdsourcing,lukes-sogaard-2018-sentiment,huang-paul-2018-examining}. Given that bag-of-word models are rarely used now, we do not perform experiments with them. Instead, we provide results with biLSTM representations initialized with random vectors for word representations. These learn only from the training data and their performance mirrors many of the  trends reported in older work. The results are shown in Table~\ref{table:rndm_summary_change}. Contrary to the results with pre-trained representations, most deterioration scores are negative, large in magnitude and statistically significant. Adaptation scores are consistent i.e.\ positive and statistically significant but have larger magnitudes than those with pre-trained representations.
Pre-training on unlabeled data injects background knowledge into models beyond the training data and has led to significant improvement on many NLP tasks. It also helps avoid or reduce the extent of temporal deterioration in models, making deployed models more (though not completely) robust to changes over time.

\begin{table}[t]
\centering
\footnotesize
\begin{tabular}{lc@{\hspace{0.25em}}c@{\hspace{0.25em}}cc@{\hspace{0.25em}}c@{\hspace{0.25em}}c@{\hspace{0.25em}}c} \toprule
& $M_s^{s+1}$ & $M_s^n$ & $M_{n-1}^n$ & $D_{t}^{a}$ & $A_{t}^{a}$ & $D_{t}^{t-1}$ & $A_{t}^{t-1}$\\ \cmidrule(lr){2-4}\cmidrule(lr){5-8}

NER & 21.8 & 10.1 & 23.9 & -6.6$^{*}$ & 6.4$^{*}$ & -2.7$^{*}$ & 3.4$^{*}$ \\
Truecasin & 89.0 & 86.0 & 88.2 & -1.5$^{*}$ & 0.7$^{\hphantom{*}}$ & -0.7$^{*}$ & 0.5$^{\hphantom{*}}$\\
Sentiment & 41.6 & 37.7 & 59.7 & -0.2$^{\hphantom{*}}$ & 9.0$^{*}$ & -0.3$^{\hphantom{*}}$ & 4.2$^{*}$ \\
Domain & 59.7 & 48.0 & 68.6 & -5.7$^{*}$ & 16.7$^{*}$ & -2.1$^{*}$ & 7.2$^{*}$ \\ \bottomrule

\end{tabular} 
\caption{Deterioration and Adaptation scores for biLSTM with randomly initialized word representations fine-tuned on gold standard data. An asterisk marks statistically significant scores.}
\label{table:rndm_summary_change}
\end{table}

\begin{table}[t]
\centering
\footnotesize
\begin{tabular}{lc@{\hspace{0.25em}}c@{\hspace{0.25em}}cc@{\hspace{0.25em}}c@{\hspace{0.25em}}c@{\hspace{0.25em}}c} \toprule
& $M_s^{s+1}$ & $M_s^n$ & $M_{n-1}^n$ & $D_{t}^{a}$ & $A_{t}^{a}$ & $D_{t}^{t-1}$ & $A_{t}^{t-1}$\\ \cmidrule(lr){2-4}\cmidrule(lr){5-8}

\multicolumn{8}{c}{NER-TTC}\\ \midrule
GloVe & 55.2 & 54.1 & 63.0 & -1.3$^{\hphantom{*}}$ & 4.1$^{*}$ & -0.1$^{\hphantom{*}}$ & 2.1$^{*}$\\
Gl+ELMo & 59.6 & 63.1 & 68.7 & 0.7$^{\hphantom{*}}$ & 1.5$^{*}$ & 1.0$^{\hphantom{*}}$ & 1.0$^{\hphantom{*}}$\\
BERT & 64.7 & 71.7 & 76.2 & 2.7$^{\hphantom{*}}$ & 1.1$^{*}$ & 2.9$^{\hphantom{*}}$ & 0.7$^{*}$\\
RoBERTa & 67.5 & 77.8 & 80.0 & 3.2$^{\hphantom{*}}$ & 1.4$^{*}$ & 3.5$^{\hphantom{*}}$ & 0.8$^{\hphantom{*}}$ \\ \midrule

\multicolumn{8}{c}{Truecasing-NYT}\\ \midrule
GloVe & 93.8 & 93.0 & 94.6 & -0.6$^{*}$ & 0.3$^{\hphantom{*}}$ & -0.2$^{*}$ & 0.3$^{\hphantom{*}}$\\
Gl+ELMo & 94.4 & 93.4 & 95.1 & -0.6$^{*}$ & 0.5$^{*}$ & -0.3$^{*}$ & 0.3$^{*}$\\
BERT & 97.2 & 94.0 & 94.6 & -1.1$^{\hphantom{*}}$ & 0.3$^{*}$ & -0.8$^{\hphantom{*}}$ & 0.2$^{*}$\\
RoBERTa & 97.5 & 94.4 & 95.6 & -1.1$^{\hphantom{*}}$ & 0.4$^{*}$ & -0.8$^{\hphantom{*}}$ & 0.2$^{*}$ \\ \midrule

\multicolumn{8}{c}{Sentiment-AR}\\ \midrule
GloVe & 44.9 & 42.8 & 64.7 & 0.8$^{\hphantom{*}}$ & 10.3$^{*}$ & 0.4$^{\hphantom{*}}$ & 4.9$^{*}$\\
Gl+ELMo & 55.3 & 57.5 & 69.1 & 2.6$^{*}$ & 5.5$^{*}$ & 1.2$^{*}$ & 2.2$^{*}$\\
BERT & 63.1 & 65.9 & 75.2 & 2.4$^{*}$ & 4.7$^{*}$ & 1.3$^{*}$ & 2.0$^{*}$\\
RoBERTa & 69.9 & 73.9 & 78.9 & 2.5$^{*}$ & 2.5$^{*}$ & 1.3$^{*}$ & 1.1$^{*}$\\ \midrule

\multicolumn{8}{c}{Domain-NYT}\\ \midrule
GloVe & 73.0 & 68.4 & 78.1 & -2.7$^{*}$ & 7.9$^{*}$ & -0.5$^{\hphantom{*}}$ & 3.6$^{*}$\\
Gl+ELMo & 77.9 & 70.7 & 82.8 & -3.9$^{*}$ & 9.4$^{*}$ & -1.0$^{\hphantom{*}}$ & 4.3$^{*}$\\
BERT & 82.7 & 74.3 & 86.2 & -4.6$^{*}$ & 9.4$^{*}$ & -1.3$^{*}$ & 4.2$^{*}$\\
RoBERTa & 84.2 & 78.2 & 86.6 & -3.7$^{*}$ & 5.8$^{*}$ & -1.1$^{*}$ & 2.9$^{*}$\\ \bottomrule

\end{tabular} 
\caption{Deterioration and Adaptation scores for models fine-tuned on gold standard data with various input representations. An asterisk marks statistically significant scores.}
\label{table:summary_change_full}
\end{table}

\section{Different Pre-trained Representations}
\label{sec:repr}

Given the variety of language representations, it is tempting to choose one for experimentation and assume that findings carry over to all. We present results using four different representations and find that this convenient assumption does not bear out in practice. We use popular representations that are likely to be used out-of-the-box.\footnote{Admittedly, input representation is an overloaded term that encompasses the model architecture, whether the representation is contextual or not, what data is used for pre-training, the overall model size, the length of the final vector representation, etc. We discuss several of these differentiating factors later.}

Both temporal model deterioration and temporal domain adaptation vary vastly across representations (Table~\ref{table:summary_change_full}). RoBERTa stands out as the representation for which results deteriorate least and for which the potential for temporal adaptation is also small. RoBERTa exhibits significant deterioration with respect to the anchor only for domain prediction; it significantly improves over time for sentiment prediction,
and changes are not significant for NER and truecasing. The improvements from temporal adaptation with respect to the anchor are statistically significant for all tasks for RoBERTa, but smaller in size compared to the improvements possible for the other representations. GloVe in contrast shows performance deterioration with respect to the anchor for three tasks (NER, truecasing and domain prediction), significant for the last two; on sentiment analysis performance of the GloVe model improves slightly, but not significantly while all other representations show significant improvements over time.

The tables also allow us to assess the impact of using a new (more recent state-of-the-art) pre-trained representation vs. annotating new training data for the same pre-trained representation. An approximate comparison can be made between two representation A and B where A is the older representation, by comparing $M_n^{n-1}$ of A i.e.\ training on $n-1$ instead of $s$ to $M_s^{n}$ of B i.e.\ still training on $s$ but using representation B. For example, consider NER with $M_s^{n}$ using GloVe as 54.1. By retraining the model with new training data, the F1 obtained is 63.0. However, by using newer representation of GloVe+ELMO, BERT and RoBERTa with the old training data, the F1 us 63.1, 71.7 and 77.8 respectively. The benefit of using new training data vs. new pre-trained representations is again highly dependent on the task and representation.

\begin{table}[t]
\centering
\footnotesize
\begin{tabular}{lrr} \toprule
\multicolumn{1}{c}{Model/Task} & \multicolumn{1}{c}{Corpus} & \multicolumn{1}{c}{Time Span}\\ \midrule
\multicolumn{3}{c}{Task Dataset}\\ \midrule
NER & TTC & 2014-2019\\
Truecasing & NYT & 1987-2004\\
Sentiment & Amazon Reviews & 2001-2018\\
Domain & NYT & 1987-2004\\ \midrule

\multicolumn{3}{c}{Pretraining Data}\\ \midrule
GloVe & Common Crawl & till 2014$^*$\\
ELMo & 1B Benchmark & till 2011$^*$\\ \cmidrule(lr){1-3}
& Wikipedia & Jan 2001-2018$^*$\\
BERT & BookCorpus & till 2015$^*$\\ \cmidrule(lr){1-3}
& Wikipedia & Jan 2001-2018$^*$\\
& BookCorpus & till 2015$^*$\\
& CC-News & Sept 2016-Feb 2019\\
& OpenWebText & till 2019$^*$\\
RoBERTa & Stories & till 2018$^*$\\ \bottomrule
\end{tabular} 

\caption{Time Span for all datasets and corpora. All corpora only include English data. $^*$ denotes that the actual time span is unknown so we note the publication date of the dataset/paper instead.}
\label{table:time_span}
\end{table}

\section{Pre-training Data Time Period}
\label{sec:time}

To perform clear experiments, one would need to control the time period of not only the task dataset but also the pre-training corpora. We report the time span for each dataset and the pre-training corpus of each model in Table~\ref{table:time_span}. For several corpora, the actual time span is unknown so we report the time of dataset/paper publication instead. This table makes it easy to spot where cleaner experimental design may be needed for future studies.

Most pre-training corpora overlap with the task dataset time span, making it hard to isolate the impact of temporal changes. BERT is trained on Wikipedia containing data from its launch in 2001 till 2018, and 11k books, spanning an unknown time period. RoBERTa uses all of the data used by BERT, and several other corpora. The pre-training data of both BERT and RoBERTa overlaps with all training and test periods of the datasets used.

We also use GloVe and ELMo representations, which do not overlap with the TTC dataset (2014-2019). GloVe was released in 2014, hence is trained on data prior to 2014. ELMo uses the 1B benchmark for pre-training which has data from WMT 2011. Yet for these two, change in model performance over time is statistically insignificant, consistent with the cases when there is overlap. Adaptation scores are higher but cannot be attributed to the lack of overlap since there is high potential for adaptation even when there is overlap for the other tasks.
GloVe and ELMo also do not overlap with a portion of the Amazon Reviews dataset (2016-2018). This is the last temporal split and is therefore used only for evaluation. Since the pre-training data does not overlap with this split, model deterioration might be expected but results on this split follow the same trend of increasing F1 with time. Table~\ref{table:amazon_test} shows the average deterioration score w.r.t.\ the anchor for all splits and the 2016-2018 split, both of which are positive. Therefore, we have at least a subset of experiments free of confounds due to pre-training time overlap. The observed trends hold across both the set of experiments with and without overlap.

\begin{table}[t]
\centering
\footnotesize
\begin{tabular}{lcc} \toprule
& All splits & Last split\\ \cmidrule(lr){2-3}
GloVe & 0.8 & 1.0\\
GloVe+ELMo & 2.6 & 3.1\\ \bottomrule
\end{tabular}
\caption{Deterioration score w.r.t.\ the anchor for Amazon Reviews averaged over all temporal test splits compared to the last temporal split which does not overlap with the pre-training data time period. Scores are positive for both.}
\label{table:amazon_test}
\end{table}

\begin{table}[t]
\centering
\footnotesize
\begin{tabular}{lrr} \toprule
\multicolumn{1}{c}{Size} & \multicolumn{1}{c}{BERT} & \multicolumn{1}{c}{RoBERTa}\\ \midrule
Large & 340M & 355M\\
Base & 110M & 125M\\
Distil-base & 65M & 82M\\ \bottomrule
\end{tabular} 
\caption{Number of parameters in the models}
\label{table:num_param}
\end{table}

\begin{table*}[t]
\centering
\footnotesize
\begin{tabular}{lc@{\hspace{0.4em}}c@{\hspace{0.4em}}cc@{\hspace{0.4em}}c@{\hspace{0.4em}}c@{\hspace{0.4em}}cc@{\hspace{0.4em}}c@{\hspace{0.4em}}cc@{\hspace{0.4em}}c@{\hspace{0.4em}}c@{\hspace{0.4em}}c} \toprule

& \multicolumn{7}{c}{BERT} & \multicolumn{7}{c}{RoBERTa}\\ \cmidrule(lr){2-4}\cmidrule(lr){5-8}\cmidrule(lr){9-11}\cmidrule(lr){12-15}

& $M_s^{s+1}$ & $M_s^n$ & $M_{n-1}^n$ & $D_{t}^{a}$ & $A_{t}^{a}$ & $D_{t}^{t-1}$ & $A_{t}^{t-1}$ & $M_s^{s+1}$ & $M_s^n$ & $M_{n-1}^n$ & $D_{t}^{a}$ & $A_{t}^{a}$ & $D_{t}^{t-1}$ & $A_{t}^{t-1}$\\ \cmidrule(lr){2-8}\cmidrule(lr){9-15}

\multicolumn{15}{c}{NER-TTC}\\ \midrule
distil-base & 59.3 & 60.0 & 69.0 & -0.4$^{\hphantom{*}}$ & 3.5$^{*}$ & 0.9$^{\hphantom{*}}$ & 1.8$^{*}$ & 60.7 & 67.2 & 70.8 & 2.1$^{\hphantom{*}}$ & 1.9$^{*}$ & 2.7$^{\hphantom{*}}$ & 0.9$^{*}$\\
base & 64.1 & 65.6 & 72.1 & 0.9$^{\hphantom{*}}$ & 2.2$^{*}$ & 1.6$^{\hphantom{*}}$ & 0.9$^{*}$ & 66.8 & 73.6 & 76.0 & 2.3$^{\hphantom{*}}$ & 0.3$^{\hphantom{*}}$ & 2.8$^{\hphantom{*}}$ & 0.4$^{\hphantom{*}}$\\
large & 64.7 & 71.7 & 76.2 & 2.7$^{\hphantom{*}}$ & 1.1$^{*}$ & 2.9$^{\hphantom{*}}$ & 0.7$^{*}$ & 67.5 & 77.8 & 80.0 & 3.2$^{\hphantom{*}}$ & 1.4$^{*}$ & 3.5$^{\hphantom{*}}$ & 0.8$^{\hphantom{*}}$\\ \midrule

\multicolumn{14}{c}{Truecasing-NYT}\\ \midrule
distil-base & 96.9 & 93.7 & 95.1 & -1.3$^{*}$ & 0.4$^{*}$ & -0.8$^{\hphantom{*}}$ & 0.3$^{*}$ & 96.4 & 93.0 & 94.4 & -1.2$^{\hphantom{*}}$ & 0.4$^{*}$ & -0.8$^{\hphantom{*}}$ & 0.3$^{*}$\\
base & 97.1 & 93.8 & 95.2 & -1.2$^{\hphantom{*}}$ & 0.4$^{*}$ & -0.8$^{\hphantom{*}}$ & 0.3$^{*}$ & 97.0 & 93.8 & 95.1 & -1.2$^{\hphantom{*}}$ & 0.4$^{*}$ & -0.8$^{\hphantom{*}}$ & 0.2$^{*}$\\
large & 97.2 & 94.0 & 94.6 & -1.1$^{\hphantom{*}}$ & 0.3$^{*}$ & -0.8$^{\hphantom{*}}$ & 0.2$^{*}$ & 97.5 & 94.4 & 95.6 & -1.1$^{\hphantom{*}}$ & 0.4$^{*}$ & -0.8$^{\hphantom{*}}$ & 0.2$^{*}$\\ \midrule

\multicolumn{14}{c}{Sentiment-AR}\\ \midrule
distil-base & 59.9 & 62.9 & 73.7 & 2.9$^{*}$ & 5.7$^{*}$ & 1.6$^{*}$ & 2.4$^{*}$ & 65.8 & 70.1 & 76.4 & 2.8$^{*}$ & 3.2$^{*}$ & 1.4$^{*}$ & 1.4$^{*}$\\
base & 63.1 & 65.9 & 75.2 & 2.4$^{*}$ & 4.7$^{*}$ & 1.3$^{*}$ & 2.0$^{*}$  & 69.9 & 73.9 & 78.9 & 2.5$^{*}$ & 2.5$^{*}$ & 1.3$^{*}$ & 1.1$^{*}$\\ \midrule

\multicolumn{14}{c}{Domain-NYT}\\ \midrule
distil-base & 81.7 & 72.9 & 85.2 & -4.9$^{*}$ & 9.8$^{*}$ & -1.4$^{\hphantom{*}}$ & 4.4$^{*}$ & 83.2 & 75.6 & 86.1 & -4.2$^{*}$ & 7.8$^{*}$ & -1.2$^{\hphantom{*}}$ & 3.6$^{*}$\\
base & 82.7 & 74.3 & 86.2 & -4.6$^{*}$ & 9.4$^{*}$ & -1.3$^{*}$ & 4.2$^{*}$ & 84.2 & 78.2 & 86.6 & -3.7$^{*}$ & 5.8$^{*}$ & -1.1$^{*}$ & 2.9$^{*}$\\ \bottomrule

\end{tabular} 
\caption{Deterioration and Adaptation scores for different model sizes with same architecture and pre-training data, fine-tuned on gold standard data. An asterisk marks statistically significant scores.}
\label{table:size_summary}
\end{table*}

Additionally, for the domain classification task, the pre-training time period overlaps with the training and evaluation years, yet we still observe considerable model deterioration. Both experiments where we do not observe deterioration despite no overlap and observe deterioration despite overlap, point to the lower impact of pre-training time period on the downstream task. Instead, these results suggest that changes in performance  are task dependent and performance is most impacted by the size of the pre-training data or the differences in model architecture.

Regardless, an important set of experiments for future work would involve pre-training the best performing model on different corpora controlled for time and compare their performance. Such an extensive set of experiments would require significant computational resources as well as time. Because of this, prior work has, like us, worked with off-the-shelf pre-trained models. For instance, \citet{Rttger2021TemporalAO} control the time period for the data used for intermediate pre-training in their experiments, but they start their experiments with BERT which is pre-trained on corpora that overlap temporally with their downstream task dataset. For future work, we emphasize the need to report the time period of any data used to support research on temporal model deterioration and temporal domain adaptation.

\section{Model Size}
\label{sec:size}


Lastly, we assess the differences in temporal effects between models with the same architecture and pre-training data\footnote{distil-RoBERTa uses less pre-training data than RoBERTa.} but different sizes. We use three versions for BERT and RoBERTa---large, base and distil-base. The distil-base models are trained via knowledge distillation from the base model \cite{sanh2019distilbert}. The number of parameters in each are reported in Table~\ref{table:num_param}. For sequence labeling (named entity recognition and truecasing), we compare all three versions. For text classification, we do not train the large model. 

Results are shown in Table~\ref{table:size_summary}. As expected, the smaller model sizes have lower F1 across all tasks, though the impact of the model size on the amount of deterioration and possible adaptation varies across task. In truecasing, there is no or little difference between the deterioration and adaptation scores across model sizes. For all other tasks, there is generally more deterioration (or less improvement for a positive score) and more scope for adaptation via retraining for smaller models. Nonetheless, the overall trend i.e.\ the direction of change in performance is consistent across model sizes.

Unlike language models which experience similar change in perplexity over time for different model sizes \cite{Lazaridou2021PitfallsOS}, we find that larger models show less deterioration (or more increase) and allow for less room for adaptation by retraining. Smaller models likely ``memorize'' less data and therefore depend more on the training data, thereby experiencing more deterioration and higher improvement via retraining.
This is further substantiated by the largest change in adaptation score with model size in the task of NER where entity memorization in pre-training may play a larger role in task performance \cite{agarwal-etal-2021-interpretability}.

\section{Temporal Adaptation without New Human Annotations}
\label{sec:temporal_adapt}

We found that model deterioration and the possibility of temporal adaptation need to be distinguished and both need to be measured. Model deterioration is task-dependent where some tasks suffer from deterioration and others do not. Regardless of whether model deterioration exists or not, for all tasks, performance can be improved by retraining on human-labeled data from a more recent time period. For tasks such as NER where the collection of new data can involve significant effort, this begets the question --- how can we perform temporal adaptation without collecting new human annotations. Here, we explore methods for this. Given human annotations for $d_s$ and a model trained on it,  we want to improve the performance of this model on $d_t$ without human annotations on $d_t$. For these experiments, we only use NER-TTC, Sentiment-AR and Domain-NYT since Truecasing-NYT showed little change in performance even when retrained with even gold-standard data.

\subsection{Continual Pre-training}
\label{sec:int_pretrain}

For the first experiment, we use domain adaptive pre-training \cite{gururangan-etal-2020-dont} on temporal splits. A pre-trained model undergoes a second pre-training on domain-specific unlabeled data before fine-tuning on task-specific labeled data. In our case, the new unabeled data is a future temporal split. However, unlike in typical domain adaptive pre-training, we only have a small amount of in-domain data. In practice, the amount of this data would depend upon how frequently one wants to retrain the model. For the experiments, we use the data from temporal split $d_t$, throwing away the gold-standard annotations. We take a pre-trained model, continue pre-training it on $d_t$, then fine-tune it on $d_s$. This is done with three random seeds and the performance is averaged over these runs. With this setup, we observe a drastic drop in performance.  We hypothesized this is because the amount of in-domain data is insufficient for stable domain adaptation. However, recent work \cite{Rttger2021TemporalAO} has shown that temporal adaptation through continual pre-training even on millions of examples has limited benefit. It should be noted that \citet{Rttger2021TemporalAO} adapt a pre-trained BERT which was pre-trained on recent data overlapping temporally with the data used for the continued pre-training. To completely disentangle the temporal effects of pre-training and assess the effective of continual pre-training, one would also need to pre-train BERT from scratch on older data.

Next, we modify the domain adaptive pre-training by adding an extra fine-tuning step. This method first performs task adaptation, followed by temporal adaptation and then again task adaptation. We start with a pre-trained model, fine-tune it on $d_s$, then pre-train it on $d_t$ and then fine-tune it again on $d_s$.  While this method does not improve performance consistently (Table~\ref{table:summary_adaptation}), it leads to significant improvement for NER and sentiment for the BERT representation, but makes no difference for domain adaptation. For RoBERTa, however, adaptation scores get worse, significantly for NER and the improvement for sentiment analysis is smaller in absolute value than for BERT.

As highlighted in the evaluation setup, multi-test set evaluation is essential for reliable results. In this experiment, if we had evaluated only on 2019 for NER-TTC (numbers omitted here), we would have concluded that this method works well but looking at the summary over different test years, one can see that the change in performance is inconsistent.


\begin{table}[t]
\centering
\footnotesize
\setlength{\tabcolsep}{3pt}
\begin{tabular}{lrrrr} \toprule
& \multicolumn{1}{c}{NER} & \multicolumn{1}{c}{Truecasing} & \multicolumn{1}{c}{Sentiment} & \multicolumn{1}{c}{Domain} \\ \cmidrule{2-5}
\multicolumn{5}{c}{BERT}\\ \midrule
Gold & 1.14$^{*}$ & 0.29$^{*}$ & {\bf 4.70}$^{*}$ & {\bf 9.37}$^{*}$\\
Pretrain & 0.84$^{*}$ & - & 1.43$^{*}$ & -0.01$^{\hphantom{*}}$\\
Self-Label & {\bf 2.27}$^{*}$ & - & 1.56$^{*}$ & 1.14$^{*}$\\ \midrule
\multicolumn{5}{c}{RoBERTa}\\ \midrule
Gold & 1.39$^{*}$ & 0.35$^{*}$ & {\bf 2.49}$^{*}$ & {\bf 5.83}$^{*}$\\
Pretrain & -0.84$^{*}$ & - & 0.25$^{*}$ & -1.34$^{\hphantom{*}}$\\
Self-Label & {\bf 1.79}$^{*}$ & - & 1.40$^{*}$ & 1.01$^{*}$\\ \bottomrule
\end{tabular} 
\caption{Adaptation scores w.r.t.\ anchor time period for different adaptation methods. Large model is used for sequence labeling and base model for text classification.}
\label{table:summary_adaptation}
\end{table}

\subsection{Self-labeling}
\label{sec:self_train}

Self-labeling has been shown to be as effective technique to detect the need to retrain a model  \cite{elsahar-galle-2019-annotate}. Here, we explore its use in temporal domain adaptation. We fine-tune a model on $d_s$, use this model to label the data $d_t$ and then use gold-standard $d_s$ and self-labelled $d_t$ to fine-tune another model. The new model is trained on $train_s$ and the full $d_t$ with $dev_s$ as the development set. $d_t$ is weakly labelled (with model predictions) and thus noisy, hence we do not extract a development set from $d_t$ for reliable evaluation. Self-labeling works consistently well, as seen in the results in Table~\ref{table:summary_adaptation}, across test years, representations and tasks\footnote{Adding new data is computationally expensive. For NER, since the amount of data is small because it required actual annotation, we could continue using the same GPU, and just the run time increased. With reviews, we had to upgrade our usage from one to two GPUs in parallel.}. Though adding self-labelled data $d_{t-1}$ does not give the highest reported performance on $d_t$, it improves performance over using just the gold-standard data $d_s$. For NER, F1 improves over using even the $d_t$ gold-standard data\footnote{For NER using BERT, we vary the amount of self-labeled new data added and observe that with 25\% of new self labeled data, adaptation score exceeds gold-standard fine-tuning (Figure \ref{fig:self_label_prop} in the appendix).} (but not over $d_s + d_t$ gold-standard data). For sentiment, F1 improves over using just gold-standard $d_s$ but not to the same level as using new gold-standard data for fine-tuning.

Lastly, we explore if continuously adding new self-labeled data further improves performance. All of $d_{s+1}$ to $d_t$ is self-labelled and added to the gold-standard $d_s$. We were able to perform this experiment only for NER since the cumulative data for reviews and domains becomes too large.
Adding more data does not improve performance but it does not decrease performance either (numbers omitted), despite the fact that the training data now comprises mainly noisy self-labelled data. More research on optimal data selection with self-labeling is needed. The right data selection may improve performance further.

\section{Experimental Design Recommendations}
\label{sec:recommend}

With this study on the impact of various factors in model training and evaluation that may confound the study of temporal effects, we recommend the following setup for experiments. We highlight the factors that can affect the findings of the study considerably and others that are less important.

\begin{enumerate}

\item Evaluate performance on the full grid of possible training and testing time periods. Variation across time-periods is considerable and choosing only one can lead to misleading conclusions about changes in performance and utility of methods.

\item Draw development data from the training year and not from the test year to ensure feasibility of the setup when used in practice.

\item Use multiple input representations since the possibility of improvement via retraining (with labeled or unlabeled data) is representation dependent and we would want an adaptation method that works consistently well.

\item Whenever possible, run experiments without overlap of the time period between the pre-training data and the task data.
This will be beneficial for clearing doubts about the reason for performance change. However, such experiments are not necessary since the observed trends seem largely unaffected by such overlap. At a minimum, report the time period for all data used (pre-training, task, external resources).

\item In case of computational constraints, use smaller models. This should not affect trends in the findings. Observed trends are similar across model sizes, even though larger models have better absolute task metrics.

\end{enumerate}

\section{Limitations and Future Work}

Work on temporal effects on a variety of tasks, domains and languages is limited by the need to collect a large amount of labeled data. We presented experiments for a range of tasks but focus only on tasks where the answer does not change with time. Other tasks such as question answering and entity linking are time-dependent and are likely to experience deterioration. Two such studies has been performed by \citet{dhingra2021time,Lazaridou2021PitfallsOS} for questions answering.

In addition, all of our experiments are on English data. Studying this for other languages, especially those with lower resources, which are most likely to experience deterioration, is harder again due to the need to collect large datasets. Though for multilingual models, one might observe the same trends as English due to transfer learning, experimental evidence will be needed in future work. Additionally, to study adaptation techniques with training on source language or source domain and evaluation on target language or target domain, one would need to match time, domain and task for both, further making such a study harder to execute. Temporal effects are hard to study but future work on different domains and languages will be beneficial, especially in light of our finding that there is not always model deterioration.

Another limitation of our work is due to the the need for large amount of computational resources needed for pre-training from scratch. \citet{Rttger2021TemporalAO} perform extensive experiments on temporal adaptation with continual pre-training but start with a pre-trained BERT which overlaps with task data. Even with the right resources, determining the timestamp for each sentence in the pre-training data is challenging. Wikipedia, a common source of pre-training data, consists of edit histories but there are frequent edits even in the same sentence. If one considers the date when the article was first added, then future data due to edits will get included. Though our experiments hint that task-pre-training data overlap may not impact the results on studies on temporal effects, a clean set of experiments with no and varying levels of overlap will be essential to understand the effect of such an overlap and motivate the selection of the pre-training data.

Finally, our analyses of statistical significance of the performance deterioration and adaptation improvement is based on the differences in performance between time periods, for scores averaged across three runs of the model. We report the minimum and maximum adaptation score across runs to account for variation across seeds. However, a single detailed test that takes in account both these variations needs to be designed carefully \cite{reimers-gurevych-2017-reporting, dror-etal-2018-hitchhikers}.
Such analysis will be able to better address questions related to whether it will be more advantageous to update the representations used for the task or to do temporal adaptation. Nevertheless, our work convincingly shows that for individual tasks and representations, deterioration either with respect to an anchor time period or for consecutive time periods is often not statistically significant. Adaptation improvements however are typically significant. This key finding will inform future work.

\section{Conclusion}

We presented exhaustive experiments to quantify the temporal effects on model performance. We outline an experimental design that allows us to draw conclusions about both temporal deterioration and the potential for temporal domain adaptation. We find that with pre-trained embeddings, temporal model deterioration is task-dependent and a model need not necessarily experience deterioration for tasks where the label correctness does not depend on time. This finding holds true regardless of whether the pre-training data time period overlaps with the task time period or not. Despite this, temporal adaptation via retraining on new gold-standard data is still beneficial. Therefore, we implemented two methods for temporal domain adaptation without labeling new data. We find that intermediate pre-training is not suitable for temporal adaptation. Self-labeling works well across tasks and representations. This finding motivates future work on how to select data to be labelled and how to maintain a reasonable size for the training data as the continual learning progresses over time.

\section*{Acknowledgments}
We thank the anonymous reviewers and the action editor, Roi Reichart, for their valuable feedback, especially the suggestion to add another task, incorporate experiments with various model sizes and perform statistical significance testing. We also thank Tracy Holloway King for her comments and careful proofreading.

\bibliography{tacl2021}
\bibliographystyle{acl_natbib}

\appendix

\section{Sentiment Classification Evaluation}

For sentiment classification, we noted the F1 for the negative class above since the positive class forms 80\% of the data and is easier to learn. Here we note the macro-F1 across both classes.

\begin{table}[h]
\centering
\footnotesize
\begin{tabular}{lc@{\hspace{0.4em}}c@{\hspace{0.4em}}cc@{\hspace{0.4em}}c@{\hspace{0.4em}}c@{\hspace{0.4em}}c} \toprule
& $M_s^{s+1}$ & $M_s^n$ & $M_{n-1}^n$ & $D_{t}^{a}$ & $A_{t}^{a}$ & $D_{t}^{t-1}$ & $A_{t}^{t-1}$\\ \cmidrule(lr){2-4}\cmidrule(lr){5-8}
GloVe & 66.6 & 67.4 & 79.2 & 1.5 & 5.5 & 0.7 & 2.6\\
Gl+ELMo & 77.2 & 75.1 & 81.9 & 2.3 & 3.3 & 1.1 & 1.4\\
BERT & 76.9 & 80.1 & 85.4 & 2.2 & 2.6 & 1.1 & 1.1\\
RoBERTa & 81.0 & 84.7 & 87.5 & 2.2 & 1.4 & 1.1 & 0.6\\ \bottomrule
\end{tabular}
\caption{Deterioration and Adaptation scores for sentiment classification using macro-F1.}
\label{table:sentiment_macro_f1}
\end{table}

\section{Hyperparameters and Design Details}

Hyperparameters are optimized for each combination of dataset and model using the oldest temporal training and development set. The same hyperparameters are then used for the remaining temporal splits for the dataset and model combination.

The biLSTM-CRF models are trained using the code from \citet{lample-etal-2016-neural}, modified to add ELMo and to perform sentence classification. BERT and RoBERTa are fine-tuned using the implementation in HuggingFace \cite{Wolf2019HuggingFacesTS}. Remaining details of hyperparameters and design choices can be found at \url{https://github.com/oagarwal/temporal-effects}.

\section{Summary Score Across Runs}

\begin{table}[h]
\centering
\footnotesize
\setlength{\tabcolsep}{4pt}
\begin{tabular}{lcccc} \toprule
& $D_{t}^{a}$ & $A_{t}^{a}$ & $D_{t}^{t-1}$ & $A_{t}^{t-1}$ \\ \cmidrule(lr){2-5}

\multicolumn{5}{c}{NER-TTC}\\ \midrule
GloVe & - & [2.5, 7.0] & - & [1.4, 3.3]\\
RoBERTa & - & [0.7, 1.9] & - & -\\ \midrule

\multicolumn{5}{c}{Truecasing-NYT}\\ \midrule
GloVe & [-0.6, -0.5] & - & [-0.2, -0.2] & -\\
RoBERTa & - & [0.3, 0.4] & - & [0.2, 0.2]\\ \midrule

\multicolumn{5}{c}{Sentiment-AR}\\ \midrule
GloVe & - & [6.9, 14.0] & - & [3.7, 6.1]\\
RoBERTa & [2.4, 2.5] & [1.8, 3.0] & [1.2, 1.4] & [0.8, 1.3]\\ \midrule

\multicolumn{5}{c}{Domain-NYT}\\ \midrule
GloVe & [-3.1, -2.1] & [6.8, 8.7] & - & [3.3, 4.0]\\
RoBERTa & [-3.9, -3.6] & [5.4, 6.0], & [-1.2, -1.1] & [2.8, 3.0]\\ \bottomrule

\end{tabular}
\caption{Minimum and maximum of summary scores across three runs for models fine-tuned on gold-standard data, for statistically significant summary metrics. Both have the same sign, showing the trends remains the same across runs.}
\label{table:summary_change_minmax}
\end{table}

\section{Self-Labeling}

For NER-TTC, self-labeling leads to better adaptation that new gold-standard data for training. For BERT, we vary the amount of new self-labeled data added to the old gold-standard data. With just 25\% of new self-labeled data added, adaptation score exceeds gold-standard fine-tuning.

\begin{figure}[t]
    \centering
    \includegraphics[width=\linewidth]{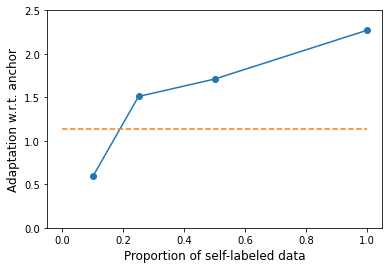}
    \caption{Adaptation score w.r.t.\ the anchor by varying the amount of self-labeled data for NER using BERT. The dashed line shows the adaptation score when just new gold-standard data is used.}
    \label{fig:self_label_prop}
\end{figure}

\end{document}